%% file: root.tex
\documentclass[letterpaper, 10 pt, conference]{ieeeconf}  % Comment this line out if you need a4paper

\IEEEoverridecommandlockouts                              % This command is only needed if 
\usepackage{times}
\usepackage{epsfig}
\usepackage{graphicx}
\usepackage{amsmath}
\usepackage{amssymb}
\usepackage{dsfont}
\usepackage{multirow}
\usepackage{url}
\usepackage{xcolor}
\usepackage{bbm}
\usepackage{subfigure}
\usepackage[font=small,labelfont=bf]{caption}
\usepackage{booktabs}
\usepackage{wrapfig}

\makeatletter
\newcommand{\eg}{\textit{e.g.}} 
\newcommand{\ie}{\textit{i.e.}}

\newcommand{\etal}{\textit{et al.}}
\makeatother

\newcommand\blfootnote[1]{%
  \begingroup
  \renewcommand\thefootnote{}\footnote{#1}%
  \addtocounter{footnote}{-1}%
  \endgroup
}

\title{\LARGE \bf
SEMI: Self-supervised Exploration via Multisensory Incongruity
}

\author{
Jianren Wang*, Ziwen Zhuang* and Hang Zhao
}

\begin{document}

\maketitle
\thispagestyle{empty}
\pagestyle{empty}
\blfootnote{Jianren Wang is with Carnegie Mellon University, jianrenw@andrew.cmu.edu; Ziwen Zhuang is with ShanghaiTech University and Shanghai Qi Zhi Institute, zhuangzw@shanghaitech.edu.cn; Hang Zhao is with Tsinghua University and Shanghai Qi Zhi Institute, hangzhao@mail.tsinghua.edu.cn; Hang Zhao is the corresponding author.}

%%%%%%%%%%%%%%%%%%%%%%%%%%%%%%%%%%%%%%%%%%%%%%%%%%%%%%%%%%%%%%%%%%%%%%%%%%%%%%%%
\begin{abstract}
Efficient exploration is a long-standing problem in reinforcement learning since extrinsic rewards are usually sparse or missing. A popular solution to this issue is to feed an agent with novelty signals as intrinsic rewards.
In this work, we introduce SEMI, a self-supervised exploration policy by incentivizing the agent to maximize a new novelty signal: multisensory incongruity, which can be measured in two aspects, perception incongruity and action incongruity. The former represents the misalignment of the multisensory inputs, while the latter represents the variance of an agent's policies under different sensory inputs.
Specifically, an alignment predictor is learned to detect whether multiple sensory inputs are aligned, the error of which is used to measure perception incongruity. A policy model takes different combinations of the multisensory observations as input, and outputs actions for exploration. The variance of actions is further used to measure action incongruity.
Using both incongruities as intrinsic rewards, SEMI allows an agent to learn skills by exploring in a self-supervised manner without any external rewards.
We further show that SEMI is compatible with extrinsic rewards and it improves sample efficiency of policy learning.
The effectiveness of SEMI is demonstrated across a variety of benchmark environments including object manipulation and audio-visual games.

\end{abstract}

%%%%%%%%%%%%%%%%%%%%%%%%%%%%%%%%%%%%%%%%%%%%%%%%%%%%%%%%%%%%%%%%%%%%%%%%%%%%%%%%
\input{sections/1-introduction.tex}

\input{sections/2-related_works.tex}

\input{sections/3-methods.tex}

\input{sections/4-experiments.tex}

\input{sections/5-conclusion.tex}

\clearpage
\bibliographystyle{ieeetr}
\bibliography{root}  % .bib

\end{document}

%% file: sections/1-introduction.tex
\section{Introduction}

% Raise the problem and challenge.
Efficient exploration is a major bottleneck in reinforcement learning problems. In many real-world scenarios, rewards extrinsic to an agent are extremely sparse or completely missing, leading to nearly random exploration of states.
A common remedy to exploration is adding intrinsic rewards, \ie, rewards automatically computed based on the agent's model of the environment. Existing formulations of intrinsic rewards include maximizing “visitation count” ~\cite{bellemare2016unifying, lopes2012exploration, poupart2006analytic} of less-frequently visited states, “curiosity”~\cite{oudeyer2009intrinsic, pathak2017curiosity, Schmidhuber:1991:PIC:116517.116542} where future prediction error is used as reward signal and “diversity rewards”~\cite{eysenbach2018diversity, lehman2011abandoning} which incentivizes diversity in the visited states. These rewards provide continuous feedback to the agent when extrinsic rewards are sparse, or even absent.
However, it is challenging to deploy these methods in practice. For “visitation count” based method, it is hard to count in a continuous space. And for “predictive model" based method, the key challenge is to model and interact with a stochastic world where multiple futures are available.

\begin{figure}
    \centering
    \includegraphics[width=0.75\linewidth]{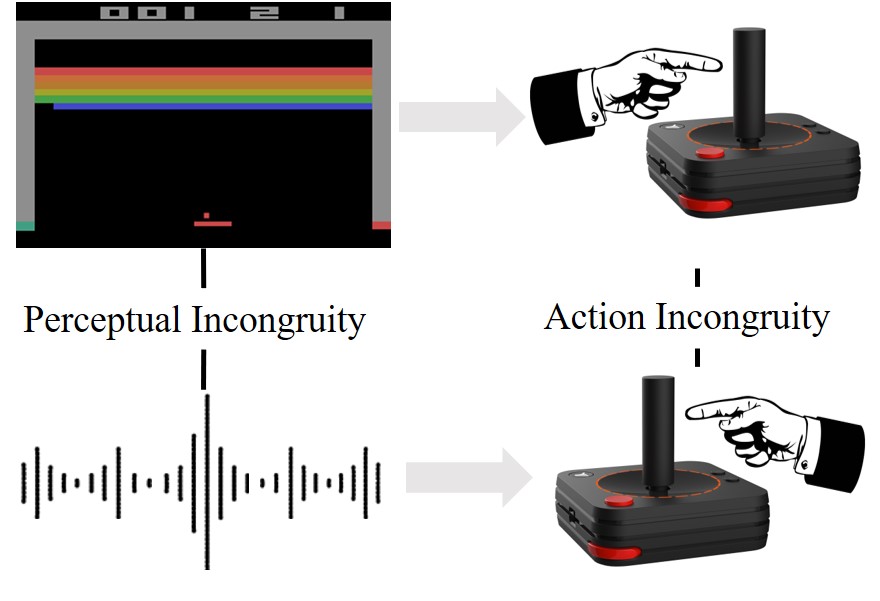}
    \caption{SEMI: a self-supervised exploration policy by incentivizing the agent to maximize multisensory incongruity, including \textit{perceptual incongruity} and \textit{action incongruity}. \textit{Perceptual incongruity} indicates the misalignment between the multisensory perceptual inputs, and \textit{action incongruity} refers to the discrepancy of actions under different perceptual inputs.}
    \label{fig:teaser}
    \vspace{-2em}
\end{figure}

% Multisensory - inspiration from humans.
As humans, we experience our world through a number of simultaneous sensory streams, such as vision, audition and touch. The novel sensory streams from different modality motivate us to explore the world and gain knowledge actively.
In modern robot design, sensors with different modalities are ubiquitous in order to augment the performance of the robot perception.
However, few exploration policies are designed around multimodal feedback for reinforcement learning agents. The difficulties are mainly reflected in two aspects: how to leverage multiple modalities with very different dimensions, frequencies and characteristics; and how to measure novelty with multimodal feedback.

% \HZ{We may want to say somewhere in the intro that, multisensory perception is becoming ubiquitous while modern robots are equipped with more sensors.}

% Our proposal
In this work, we introduce \textbf{SEMI}, a self-supervised exploration method by incentivizing the agent to maximize multisensory incongruity, including \textit{perceptual incongruity} and \textit{action incongruity}, as shown in Figure~\ref{fig:teaser}.

% Perceptual incongruity.
\textit{Perceptual incongruity} is defined as the misalignment between multisensory inputs.
As humans, the coincidence of senses gives us strong evidence that they were generated by a common, underlying event~\cite{sekuler1997sound}, since it is unlikely that they co-occurred across multiple modalities merely by chance. Thus, the misalignment or incongruity between multisensory streams can be used as a strong signal of novelty. Researches in psychology suggested that this incongruity can attract human's attention and trigger further exploration~\cite{berlyne1963novelty,dember1957analysis}, which has been widely used in product design~\cite{ludden2012beyond, ludden2008surprise}.
In SEMI, we use such novelty to guide robot exploration. Specifically, an alignment predictor is trained to detect misalignment between multisensory inputs. The model observes raw sensory streams — some of which are paired, and some have been shuffled — and we task it with distinguishing between the two. This challenging task forces the model to fuse information from multiple modalities and meanwhile learn a useful feature representation. The prediction error of the sensor fusion model serves as a metric of perceptual incongruity, which is further used as an intrinsic reward to guide the agent's exploration.

% Action incongruity.
\textit{Action incongruity} is defined as the discrepancy of an agent's decisions when it perceives different senses of the same underlying event. 
This is inspired from the fact that humans are able to integrate multimodal sensory information in a near-optimal manner for decision making~\cite{angelaki2009multisensory, ma2008linking}, and are even robust to the loss of some senses~\cite{hoover2012sensory, kolarik2014summary}. Sensory compensation empowers humans to make similar decisions when different senses are used~\cite{cohen1997functional, bavelier2002cross, lee2001cross}.
% Thus, the incongruity of decisions which are made under different combinations of senses is also a signal of novelty for robots, and can be utilized for exploration.
% {\color{red}Lack of sensory inputs might lead to complete different action in a non-robust multi-modal policy. Thus, with a multi-modal policy trained with combinations of multimodal sensory input can be utilized to produce a signal of novelty for robots in order to perform exploration.}
In SEMI, a policy network is learned with multi-modal dropout during multisensory fusion. Concretely, we randomly drop one or several modalities during multisensory fusion to imitate loss of senses. The variance of actions suggested by the policy network under different dropout states is used to measure action incongruity, which is also used as an intrinsic reward for better exploration.
 
SEMI is evaluated in two challenging scenarios: object manipulation (vision and depth) and audio-visual games (Gym Retro). We show that SEMI outperforms ``predictive model" based exploration policy by a large margin in both scenarios.

The contributions of this paper can be summarized as follows. Inspired by psychology, we propose SEMI, a novel self-supervised exploration policy through discovering multisensory perceptual and action incongruity; SEMI enables agents to learn compact multimodal representation from hard examples; we demonstrate the efficacy of this formulation across a variety of benchmark environments including object manipulation and audio-visual games; furthermore, we show that SEMI is complementary to other intrinsic and extrinsic rewards.

%% file: sections/2-related_works.tex
\section{Related Works}

\paragraph{Explore with Intrinsic Rewards.}

Consider an agent that sees an observation, takes an action and transitions to the next state. We aim to incentivize this agent with a reward relating to how informative the transition was, so that the agent can explore the complicated environment more efficiently. 
One simple approach to encourage exploration is to use state visitation counts~\cite{bellemare2016unifying, fu2017ex2, tang2017exploration}, where one maximizes visits on less frequent states. However, counting in the continuous space is usually challenging.
Recently a more popular line of works are using prediction error~\cite{pathak2017curiosity,Schmidhuber:1991:PIC:116517.116542,burda2018large,burda2018exploration}, prediction uncertainty~\cite{houthooft2016vime, osband2016deep}, or improvement~\cite{lopes2012exploration} of a forward dynamics or value model as intrinsic rewards. As a result, the agent is driven to reach regions of the environment that are difficult to reason with the current model.
Our proposed method also follows these works, but instead studies the exploration problem in a multisensory perception setting, which is becoming more common for modern robots.

A concurrent work from Dean~\etal~\cite{dean2020see} has also demonstrated the effectiveness of using multisensory signals as intrinsic rewards. Specifically, they focus on the association of audio and visual signals as intrinsic rewards for reinforcement learning exploration. Different from them, our multisensory incongruity contains both perceptual incongruity and action incongruity.

\paragraph{Multimodal Self-supervised Learning.}
Self-supervised methods learn features by training a model to solve a pretext task derived from the input data itself, without human labeling. A variety of pretext tasks have been proposed to learn representations from different modalities. Several works leverage the natural correspondence~\cite{arandjelovic2017look, patrick2020multi} and synchronization~\cite{owens2018audio, lee2019making} between the audio or tactile and RGB streams to learn representations.
% {\color{red}
In our work, we also used the natural correspondence and synchronization of different modalities and we proposed the novel method to utilize these relations and compute intrinsic reward signal for the exploration algorithm.
% }
% Others use a modality distillation framework to learn video~\cite{owens2016ambient} and sound~\cite{aytar2016soundnet} representations.
Recent works have also found that multi-modal learning can lead to more robust representations as they can partly account for the different learning speeds of the different modalities~\cite{alwassel2019self}.

\begin{figure*}[t]
  \centering
  \includegraphics[width=0.9\textwidth]{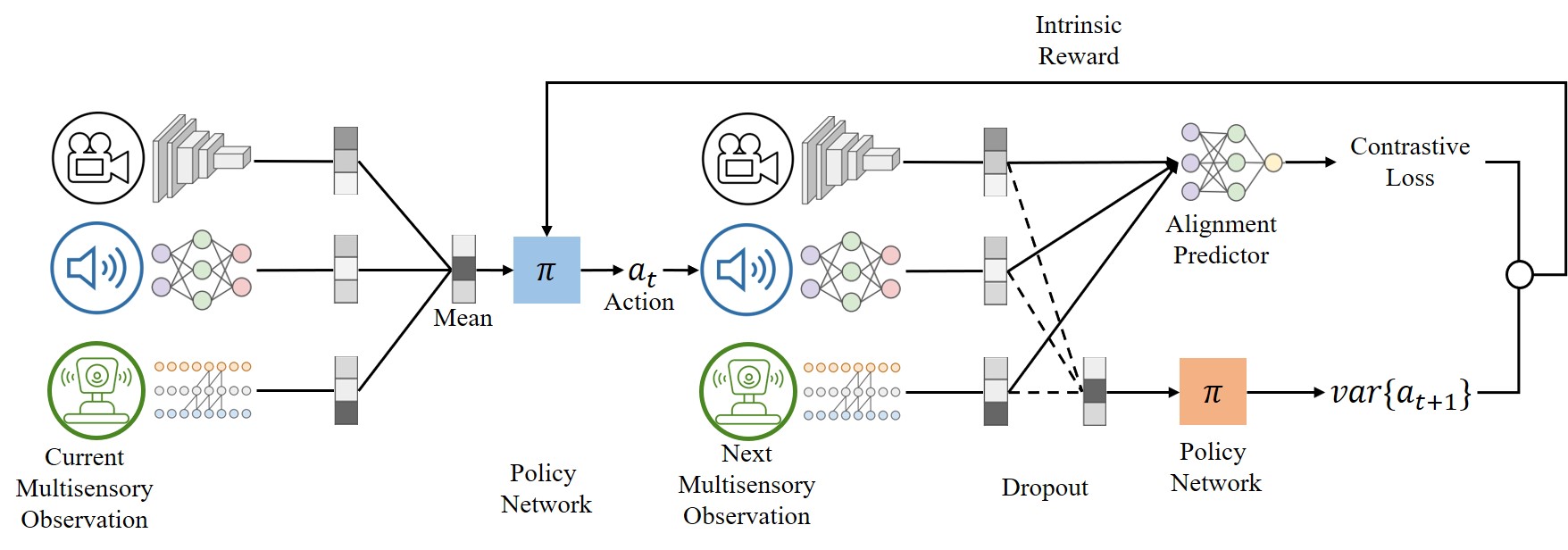}
  \caption{SEMI pipeline overview: at time step $t$, an agent takes action $a_t$ given a multisensory observation $O_t$ as input and ends up in a new state. The multisensory fusion model takes a new observation $O_{t+1}$ as input and predicts whether these sensory inputs are aligned. The prediction loss is used as the measure of perceptual incongruity. The variance of actions suggested by the policy network given different combination of multisensory inputs is used to measure action incongruity.
  Both incongruities are used as intrinsic rewards to train the policy $\pi$.}
  \label{fig:pipeline}
\end{figure*}

\paragraph{Noise-contrastive Estimation.}
Noise-contrastive estimation~\cite{hadsell2006dimensionality,gutmann2010noise,mnih2013learning}
measures the compatibility between sample pairs in a representational space and is at the core of several recent works on unsupervised feature learning~\cite{tian2019contrastive,he2019momentum,hadsell2006dimensionality,patrick2020multi,chen2020simple}.
It reduces a density estimation problem into a simpler probabilistic classification problem, circumventing the need to design handcrafted tasks in the raw signal space.
Contrastive learning has recently been shown to yield good performance for image and video representation learning~\cite{oord2018representation,he2019momentum,han2019video}. Prominently, Chen~\etal~\cite{chen2020simple} demonstrated that proper combination of data augmentation strategies and noise-contrastive re-identification achieves superior unsupervised learning results. 
% {\color{red}
In our work, we build the positive-negative pair of data augmentation by utilizing the temporal correspondence throughout the experience of the agent.
% }

%% file: sections/3-methods.tex
\section{Method}

% \subsection{Formulation}
SEMI is a self-supervised exploration policy that incentivizes agents to maximize multisensory incongruities, which we formulate as two aspects: perceptual incongruity (Section~\ref{sec:method_perceptual}) and action incongruity (Section~\ref{sec:method_action}). Both incongruities are fed to the agent as intrinsic rewards to encourage its exploration.
Figure~\ref{fig:pipeline} gives an overview of the pipeline of SEMI, and we will detail each sub-module in the following.
%\HZ{TO BE MOVED AROUND: SEMI is composed of two sub-modules: an alignment predictor that generates a perceptual incongruity reward $r_t^p$, and a policy network that generates an action together with an action incongruity reward $r_t^a$. The alignment predictor observes multiple raw sensory streams, and detects the misalignment between them. We use the prediction error as a measure of perceptual incongruity. The policy network also observes multisensory inputs. The variance of actions suggested by the policy network given different modalities is used to measure action incongruity.}

\paragraph{Notation.}
Given an agent's current observation $O_t$ at time $t$, our goal is to generate intrinsic curiosity reward $r_t$ so that the agent learns a policy $\pi$ to explore unknown and difficult environment.
In this paper, we focus on the multisensory setting, where the agent observes a set of perceptual inputs $O_t = \{o_t^1, o_t^2, ..., o_t^M\}$, where $M$ is the number of modalities, which could represent vision, audio, touch, \textit{etc}.
By executing an action $a_t$ produced by the policy, the agent further observes the next state, which we denote as $O_{t+1}=\{o_{t+1}^1, o_{t+1}^2, ..., o_{t+1}^M\}$.

\subsection{Multisensory Perceptual Incongruity}
\label{sec:method_perceptual}

The synchrony of multiple senses is a fundamental property of natural event perception. We humans are extremely sensitive to the incongruity between these senses, which is a strong signal of novelty. For example, if a common object makes an uncommon sound, we are motivated to further interact with this object to gain better knowledge about it. 
Inspired by this observation, we aim to use such novel association signals as curiosity to drive an RL agent to explore unfamiliar states. 

To guide an agent to explore novel states, we propose an alignment predictor to discover the perceptual incongruity.
Alignment prediction can take various forms, one possible design is to predict one sensory stream from other streams. For example, we could generate sounds from a corresponding visual input, or generate images from its sounds. However, generating data in the raw signal space is proved to be challenging, since (1) it does not handle the cases of multiple possible targets, (2) it suffers from overfitting to trivial details or noises~\cite{pathak2017curiosity}.

A better idea is to predict the compatibility of multisensory streams in the latent space. Along the idea of contrastive learning~\cite{oord2018representation,chen2020simple}, our design of alignment predictor directly maximizes the agreement between different modalities of the same event. This is achieved by predicting positive (aligned) modality streams from negative ones via a contrastive loss penalty in the latent space.
The predicted alignment score can then be used as an indicator of perceptual incongruity.

% For perceptual incongruity, an alignment predictor is trained to detect misalignment between multisensory inputs. The model observes raw sensory streams — some of which are paired, and some have been shuffled — and we task it with distinguishing between the two. This challenging task forces the model to fuse information from multiple modalities and meanwhile learn a useful feature representation. The prediction error of the sensor fusion model serves as a metric of perceptual incongruity, which is further used as an intrinsic reward to guide the agent's exploration.

Concretely, the alignment predictor comprises the following two major components. 

$\bullet$ A set of neural network base encoders $(f_1(\cdot), ..., f_M(\cdot))$ that extracts representation vectors from each modality. Our framework is agnostic to the choices of neural network architectures. In the following experiments, we use a 2D ConvNet to extract RGB visual features, another 2D ConvNet to obtain depth features, and a Short Time Fourier Transform (STFT) followed by a 1D ConvNet to extract the audio features.

$\bullet$ A contrastive loss function defined for a contrastive learning. Given one sensory stream $o^j$ from a multisensory observation $O = \{o^i\}|_{i=1,...,M}$ (we omit time $t$ in the following for brevity), we define the other $M-1$ simultaneous sensation streams $\{o^i\}|_{i\neq j}$ as positive examples. In a minibatch of $N$ observations, there are $M \times (N - 1)$ sensory streams from other modalities, which can be used to build misaligned examples. The contrastive prediction task aims to identify aligned sensory streams from these misaligned examples. 

The similarity of a pair of multimodal observation $(o^i,o^j)$ are measured by the cosine distance, \ie
\vspace{-3pt}
\begin{equation}
\text{sim}(o^i,o^j) = \cos(\mathbf{f_i, f_j}) = \frac{\mathbf{f^T_i} \cdot \mathbf{f_j}}{||\mathbf{f_i}|| \cdot ||\mathbf{f_j}||},
\end{equation}
where $\mathbf{f_i}=f_i(o^i), \mathbf{f_j}=f_j(o^j)$ are features from different modalities.
Then the contrastive loss function for a pair of positive observation $(o^i_k,o^j_k)$ is defined as 
\vspace{-3pt}
\begin{equation}
\mathcal{L}(o^i_k,o^j_k) = -\text{log} \frac{\text{exp}(\text{sim}(o^i_k,o^j_k)/\tau)}{\sum_{\substack{n=1}}^{N} \sum_{\substack{m=1}}^{M}\text{exp}(\text{sim}(o^i_k,o_n^m)/\tau)},
\end{equation}
where $\tau$ denotes a temperature parameter. 

The \textit{multisensory perceptual incongruity} of an observation $O_k$ is then defined numerically as the sum of losses of all possible multisensory pairs from the same timestep, which can be used as an intrinsic reward $r^p = \sum_{\substack{i=1}}^{M} \sum_{\substack{j=i+1}}^{M} \mathcal{L}(o^i_k,o^j_k)$.

\subsection{Multisensory Action Incongruity}
\label{sec:method_action}

Congruity in actions is inspired from the fact that human perception is robust to the partly loss of senses, and humans have an exceptional ability to compensate for the loss with other senses.
For example, an experienced driver can predict if cars are coming up from nearby lanes just from sound noise, without turning his/her head to look.
% An example of novel scenario is that the driver looks at the nearby lanes after hearing the sound of a car, but does not see anything. 
If we make different decisions with different sensory inputs, it suggests we have low confidence of the event we experienced, \eg an inexperienced driver might change lane recklessly without a good understanding of the distance of cars from the sound noise.
Inspired by the above observation, we further aim to use the action incongruity as an indicator of novelty in RL exploration.

Here we implement the action incongruity via drop of senses. Proposed by Srivastava~\etal~\cite{srivastava2014dropout}, \textit{dropout} has been widely used to prevent neural networks from overfitting~\cite{lecun2015deep, Huang_2017_CVPR}. Gal~\etal~\cite{gal2016dropout,NIPS2017_7141} further cast dropout training in deep neural networks as approximate Bayesian inference in deep Gaussian processes, which offers a mathematically grounded framework to reason about model uncertainty.

We adopt a similar approach by taking a sensory-wise dropout strategy during sensor fusion for the policy network. Then multisensory action incongruity is defined as the divergence of actions suggested by the policy network given different combinations of multisensory observations.

%Similarly, $O = \{o^i\}|_{i=1,...,M}$ represents the multimodal observation, and $\mathbf{f_i}=f_i(o^i)$ is the encoded feature for each modality.
Specifically, we combine features of different modalities with dropout to obtain a fused perceptual feature $z$,

\begin{equation}
    z = \frac{1}{\sum_{i=1}^M\mathds{1}^i}(\sum_{i=1}^M\mathds{1}^i \mathbf{f_i})
\label{eq:fusion}
\end{equation}
where $\mathds{1}^i \in \{0,1\}$ indicates the existence of $\mathbf{f_i}$. Apparently, different combinations of $\mathds{1}^i$ will lead to different $z$. We collect the action outputs from the policy network $\pi_r$ given all possible inputs $z$'s ($2^M-1$ possible inputs in total), and define the variance of these actions as the \textit{multisensory action incongruity}. The action incongruity is further used as an intrinsic reward $r^a$ for exploration,
\begin{equation}
    r^a = \frac{1}{2^M-1} \sum_{k=1}^{2^M-1} ||\pi_r(z^k)-\frac{1}{2^M-1}\sum_{k=1}^{2^M-1}\pi_r(z^k)||_2^2.
\label{eq:action}
\end{equation}

\subsection{Multisensory Incongruities as Intrinsic Rewards}

To summarize, we use both multisensory perceptual incongruity and multisensory action incongruity as intrinsic rewards. 
It is worth noting that the policy network $\pi_r$ used to calculate intrinsic reward $r^a_t$ is different from that used for exploration $\pi$. Inspired by Double Q-learning~\cite{van2016deep} and Dual Policy Iteration~\cite{sun2018dual}, $\pi_r$, with parameters $\theta$ being the same as $\pi$ except that its parameters are copied every $\tau$ steps from the $\pi$. This simple strategy not only reduces the observed overestimations, but also leads to better convergence.

At time step $t$, the agent takes action $a_t$ given multisensory observation $O_t$ with modality dropout as input and receives a new observation $O_{t+1}$ and intrinsic reward in calculated as $r_t = r^p_t + \gamma \times r^a_t$, where $\gamma$ is a weight factor. The agent is optimized using PPO~\cite{schulman2017proximal} to maximize the expected reward according to

\begin{equation}
\max_{\theta} \mathbf{E}_{\pi(O_t;\theta)}(\sum_t r_t).
\end{equation}

%% file: sections/4-experiments.tex
\section{Experiments}

We evaluate the performance of SEMI in two environments, \textit{OpenAI Robotics} and \textit{Atari}. Three settings are considered and discussed: exploration with multisensory incongruity only (Section~\ref{sec:intrisic_only}), combining multisensory incongruity with extrinsic reward (Section~\ref{sec:intrisic_extrinsic}), and combining multisensory incongruity with other intrinsic rewards (Section~\ref{sec:intrisic_other}).

\subsection{Exploration via Multisensory Incongruity}
\label{sec:intrisic_only}

\subsubsection{Environment and Setting}

\paragraph{OpenAI Robotics} We evaluate our method on OpenAI Robotics~\cite{plappert2018multi}, where robot receives RGB image and Depth image as two modalities, and controls the gripper Cartesian movement, gripper rotation as well as gripper open or close.
\paragraph{Atari} We also evaluate our method on Atari games, where vision and audio are considered as multi-modal inputs. We use Gym Retro~\cite{nichol2018gotta} in order to access game audio.

Further details for the two evaluation environments are described in the supplementary materials.

\subsubsection{Training Details}
\label{section:training}

In general, we used 5 convolutional layers to extract RGB features, a similar network to extract depth features or 5 consecutive frames channel-wise spectrum to represent audio feature. We used a 4-layer multi-layer perceptron (MLP) as our policy network and used PPO to maximized the intrinsic reward with an Adam Optimizer. During training, all rewards that are collected in trajectories will be replaced or added by intrinsic reward. Further details in training the agent in OpenAI Robotics and Atari Games are described in supplementary materials.

\subsubsection{Results}

\paragraph{OpenAI Robotics}
Table~\ref{tab:robot} shows the exploration performance of object manipulation using the multisensory incongruity, which are measured by the frequency at which our agent interacts (\ie, touches) with the object (\ie. interaction rate). The interaction rate is defined as \textit{\#trials robot interact with object/\#total trials}.

We evaluate two different versions of our method. We first use only the multisensory perceptual incongruity as our intrinsic reward, as described in Section~\ref{sec:method_perceptual}. Second, we use both multisensory perceptual incongruity and multisensory action incongruity as our intrinsic reward.

% \begin{itemize}[leftmargin=10pt]
%     \item \textit{SEMI (P)}:
%     \item \textit{SEMI (PA)}: 
% \end{itemize}

We compare SEMI to Curiosity~\cite{pathak2017curiosity, burda2018large} and Disagreement~\cite{pathak2019self} as our baselines. Also, we compared with a random policy as a sanity check, which samples its action uniformly from the action space.

% \begin{itemize}[leftmargin=10pt]
%     \item \textit{Curiosity}: Proposed by~\cite{pathak2017curiosity, burda2018large}, the exploration policy is trained jointly with a predictive model. The predictive model is trained to predict the next state from current state and action, and the error of which is used as intrinsic reward. The intuition is unpredictable situations are more likely novel and therefore ones the agent should explore.
%     \item \textit{Random}: As a sanity check, we propose to use a random policy, which moves randomly.
%     \item \textit{Disagreement}: Proposed by~\cite{pathak2019self}, they first train an ensemble of dynamics models and incentivize the agent to explore such that the disagreement of those ensembles is maximized.
% \end{itemize}

\begin{table*}[ht!]
\centering
\begin{tabular}{c | c | p{2.3cm} | p{2.2cm} | p{2.3cm}}
\toprule
\multicolumn{2}{c|}{Exploration Strategy}            & Interaction Rate (1 objects)  & Convergence Iteration & Interaction Rate (1 of 3 objects) \\ \hline
\multirow{5}{*}{Uni-IR}                 & Curiosity         & 2.7\%     & 25    & 8.3\% \\ 
                                        & Random            & 8.4\%     & 0     & 22.6\% \\
                                        & Disagreement      & 26.3\%    & 23    & 64.3\% \\
                                        & SEMI (P)          & 30.5\%    & 20    & 81.4\% \\
                                        & SEMI (PA)         & 34.4\%    & 33    & 82.1\% \\ \hline
\multirow{3}{*}{Multi-IR}               & Curiosity + SEMI (PA)     & 35.8\%    & 36 & 83.3\% \\
                                        & Disagreement + SEMI (PA)  & 37.1\%    & 35 & 83.8\% \\
\bottomrule
\end{tabular}
\caption{We measure the exploration quality by evaluating the object interaction frequency of the agent trained with different intrinsic rewards (Row 1-5) and a combination of intrinsic rewards (Row 6-7).}
\label{tab:robot}
\end{table*}

As shown in Table~\ref{tab:robot}, our method outperforms all of these baselines. The method of Disagreement~\cite{pathak2019self} has a performance close to that of our method.
% In Figure~\ref{fig:manipulation}, we show examples where our method interacts with objects, whereas the baseline Random policy fails. 

We perform an ablation analysis to quantify the performance of each component of our system (4th and 5th row in Table~\ref{tab:robot}). We see that both multisensory perceptual incongruity and multisensory action incongruity contribute to the robot exploration. 

\begin{figure*}[ht!]
  \centering
  \includegraphics[width=0.95\textwidth]{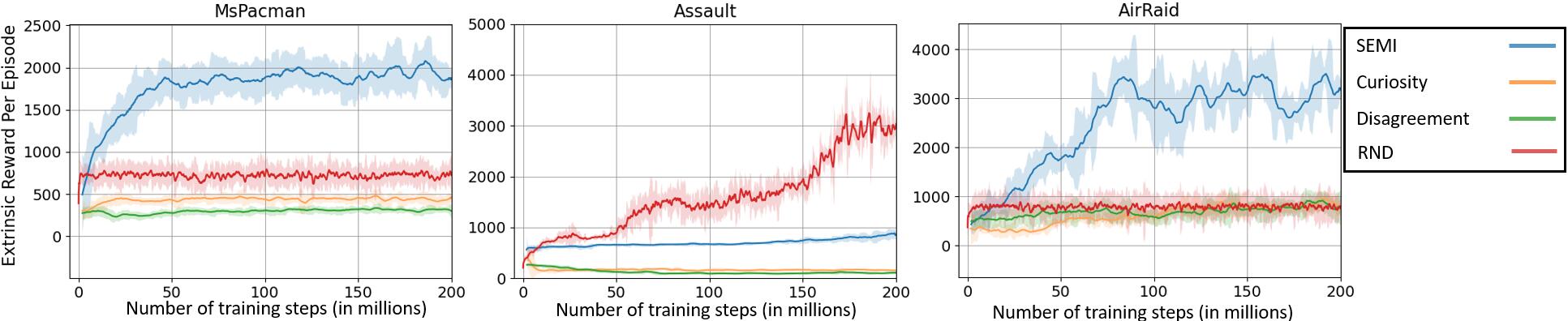}
  \caption{We compare different intrinsic reward formulations across different Atari games. We run three independent runs of each algorithm and show the mean extrinsic reward during training. SEMI far outperforms curiosity-based baseline and disagreement-based baseline, and also learns more efficiently.}
  \label{fig:atari_in}
\end{figure*}

\paragraph{Atari}
We also test out method in Atari MsPacman, Assault, AirRaid, Alien, Space Invaders, Breakout, and Beam Rider. Figure~\ref{fig:atari_in} shows the extrinsic reward of some Atari games during exploration with SEMI in comparison of intrinsic reward via RND, Curiosity and Disagreement. It should be pointed that during training the agent only has access to the intrinsic reward. As illustrated in Figure~\ref{fig:atari_in}, our method converges faster and achieves better performances comparing with all baseline methods.
% This result depicts that SEMI can explore the environment more efficiently and more thoroughly.
The reason is that audio signals are always triggered by significant events (\textit{e.g.} eating pellets).
% {\color{red}
The highly temporal-aligned signal from different sensory modalities can be explicitly utilized when we are computing the contrastive loss function.
% }
Thus, the multisensory incongruity is more indicative compared with curiosity and disagreement baselines, which are influenced by the stochasticity of the environments. 

\begin{figure*}[ht!]
  \begin{center}
    \includegraphics[width=0.9\textwidth]{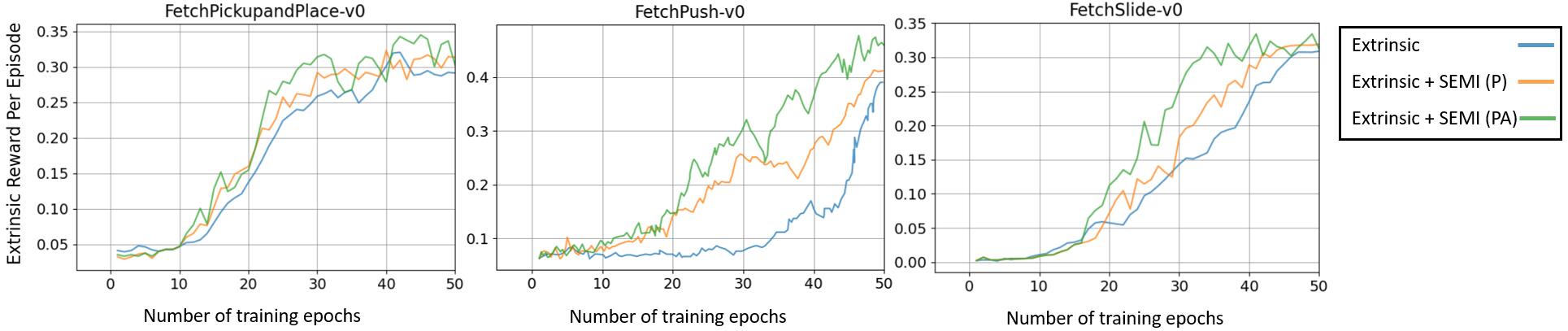}
  \end{center}
  \caption{Comparison of episodic extrinsic rewards of the agent trained with and without multisensory incongruity in the Fetch-Pushing task. Training with SEMI significantly improves the learning efficiency  compared with training with only sparse extrinsic reward.}
  \label{fig:robot_ex}
\end{figure*}

% In the meantime, the Breakout environment is deterministic, which makes it easier to learn a predictive model. And in Beam Rider, the noisy background music keeps appearing whenever the agent is making a move. These also provide no information for the multisensory incongruity module to neither learn the association nor guide the RL algorithm to learn better policy. 

\subsection{Combining with Extrinsic Reward}
\label{sec:intrisic_extrinsic}
Considering our proposed intrinsic rewards are designed to guide the agent to explore the environment, will they allow the agent to explore well when the extrinsic reward is sparse?
To verify this, we conduct additional experiments in OpenAI Robotics and Atari Games. While the network architecture and training schema are exactly the same as Section~\ref{sec:intrisic_only}, we use the sum of SEMI and extrinsic rewards as training signal,

\begin{equation}
\begin{aligned}
    R_t = r_t + \beta \times r^{(e)}_t
\label{eq:atari_ex}
\end{aligned}
\end{equation}
where $r_t^{(e)}$ is the external reward provided by the environment. We set $\beta$ to $1$ in all the experiments.

\begin{figure*}[ht!]
  \begin{center}
      \includegraphics[width=0.9\textwidth]{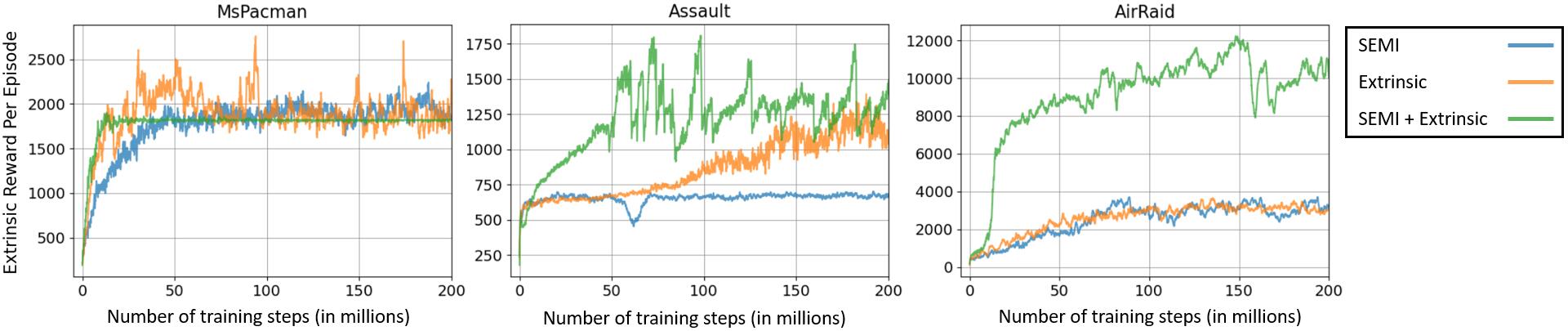}
  \end{center}
  \caption{Comparing the performance of the agent trained with multisensory incongruity against without using intrinsic rewards across different Atari games. We run three independent runs of each algorithm and show the mean extrinsic reward during training. Training with SEMI always leads to a faster convergence.}
  \label{fig:atari_ex}
\end{figure*}

Figure~\ref{fig:robot_ex} shows the episodic extrinsic reward of the FetchPush, Pickup and Place, FetchSlide task training with multisensory incongruity against without using intrinsic rewards. The extrinsic rewards are sparse and binary: The agent obtains a reward of $0$ if the goal has been achieved (within some task-specific tolerance) and $-1$ otherwise. Training with SEMI significantly improves the learning efficiency compared with training with only sparse extrinsic reward. By adding action variance in SEMI paradigm (PA), the performance improves further when policy model learns meaningful mapping from observation to action. 

We also tested the experiments on Atari MsPacman, AirRaid, Assault, and Alien. Figure~\ref{fig:atari_ex} shows the effectiveness of our method for efficient exploration in these Games. Training with SEMI always leads to a faster convergence, which indicates that it is able to speed up exploring the environments. Besides, the final performance of the agent does not deteriorate with faster convergence, showing the compatibility of SEMI with any extrinsic rewards. 

\subsection{Combining with Other Intrinsic Rewards}
\label{sec:intrisic_other}

We further show that exploration via multisensory incongruity is complementary to some other self-supervised exploration methods, \eg~prediction-based curiosity. To demonstrate this, we simply sum the multisensory incongruity with other intrinsic rewards, and use it to train the agents.
We evaluate this setup in both OpenAI Robotics and Atari Games. The network architecture and training schema are exactly the same as mentioned in Section~\ref{sec:intrisic_only}. 

\begin{figure*}[ht!]
  \begin{center}
    \includegraphics[width=0.9\textwidth]{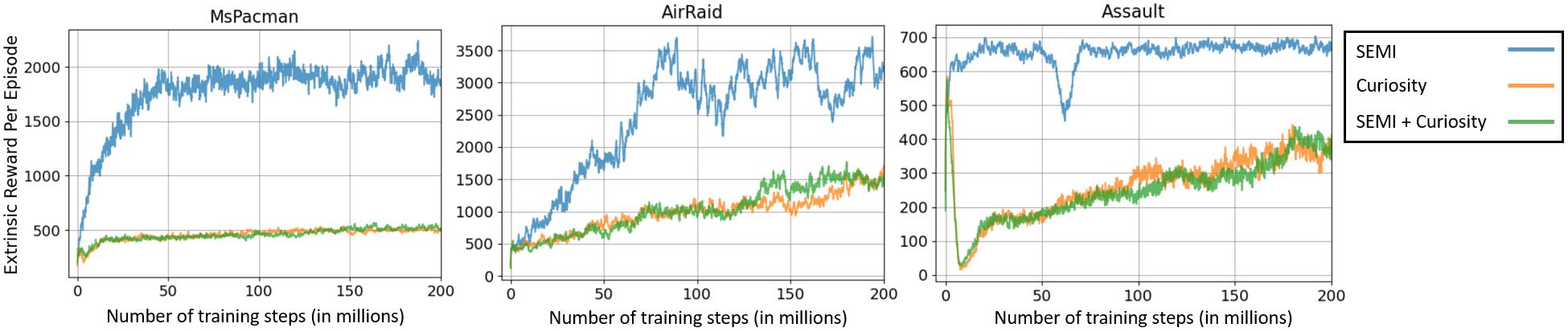}
  \end{center}
  \caption{Comparing the performance of the agent trained with multisensory incongruity joint with other intrinsic rewards against trained with multisenory incongruity alone across different Atari games. We run three independent runs of each algorithm and show the mean extrinsic reward during training.}
  \label{fig:atari_ins}
\end{figure*}

Table~\ref{tab:robot} (5th and 6th row) shows the interaction rate of object manipulation during exploration with a combination of intrinsic rewards, which sums two intrinsic reward directly as the total intrinsic reward for the agent.
% \begin{itemize}
%     \item \textit{SEMI + Curiosity}: The agent receives a combination of intrinsic rewards: we sum the losses from multisensory incongruity and visual prediction error.
%     \item \textit{SEMI + Disagreement}: The agent receives a combination of intrinsic rewards: we sum the losses from multisensory incongruity and the disagreement of dynamics model ensembles.
% \end{itemize}
The agent maximizing the sum of multiple intrinsic rewards explores better than an agent maximizing single intrinsic rewards, which shows that SEMI is complementary to many existing intrinsic rewards. 

Similarly, We combine our method with other intrinsic rewards on MaPacman, Assault and AirRaid. Figure~\ref{fig:atari_ins} shows the extrinsic reward of Atari during exploration with a combination of intrinsic rewards. On environment such as AirRaid, the extrinsic reward converge significantly faster than trained with only visual prediction or SEMI method. But on environments like Alien and Space Invaders, the performance does not improve compared to the visual prediction baselines no matter whether multisensory incongruity outperforms curiosity. Since it is unclear how the intrinsic rewards will affect each other when trained jointly, it is possible that optimizing some rewards can bring negative impacts on the others, which will lead to a worse exploration efficiency.

\subsection{Failure Cases}

While SEMI generally shows improvement in exploring the environment and is compatible with training with extrinsic reward, there are still some Atari environments where it does not improve exploration efficiency. We dig into the games and analyze the feature of them to explain why these environments lead to failures.

\paragraph{The game presents constant sound patterns} 
For example in Beam Rider, there is a fixed background sound whenever the agent makes a move. Thus, the multisensory incongruity method will not learn useful patterns to distinguish the incongruity even in the basic situations, therefore the agent cannot learn from any meaningful intrinsic reward signal.

\paragraph{The game shows trivial multisensory association}
In environments like Breakout, the audio is almost the same when the agent is interacting with the environment, \textit{i.e.} the sound in Breakout only indicates the ball is making contact with objects in the scene. The multisensory incongruity module could easily distinguish the incongruity in almost all cases in the game. A newly reached game situation will not lead to high intrinsic rewards. Therefore, multisensory incongruity method cannot motivate the agent to explore unseen situations.

%% file: sections/5-conclusion.tex
\section{Conclusion}

In conclusion, we proposed SEMI, a self-supervised exploration strategy by incentivizing the agent to maximize multisensory incongruity. We showed that through the use of multisensory perceptual incongruity and multisensory action incongruity, our learned policy can explore the environment efficiently. We also showed the compatibility of our proposed method with extrinsic rewards and other intrinsic rewards.

We hope that our work paves the way towards to a direction for intelligent agents to continually develop knowledge and acquire new skills from multisensory observations without human supervision.